%% file: main.tex
\documentclass[]{style/ceurart}
\usepackage{tcolorbox}
\tcbuselibrary{breakable}
\tcbuselibrary{listings} 
\begin{document}

\copyrightyear{2025}
\copyrightclause{Copyright for this paper by its authors.
  Use permitted under Creative Commons License Attribution 4.0
  International (CC BY 4.0).}

\conference{CLEF 2025: Working Notes, 9 -- 12 September 2025, Madrid, Spain}

\title{Transfer Learning and Mixup for Fine-Grained Few-Shot Fungi Classification}

\author[1]{Jason Kahei Tam}[%
    orcid=0009-0005-0853-0414,
    email=jtam30@gatech.edu,
]
\cormark[1]

\author[1]{Murilo Gustineli}[%
    orcid=0009-0003-9818-496X,
    email=murilogustineli@gatech.edu,
]
\cormark[1]

\author[1]{Anthony Miyaguchi}[%
    orcid=0000-0002-9165-8718,
    email=acmiyaguchi@gatech.edu,
]
\cormark[1]

\address[1]{Georgia Institute of Technology, North Ave NW, Atlanta, GA 30332}
\cortext[1]{Corresponding author.}

\begin{abstract}
    Accurate identification of fungi species presents a unique challenge in computer vision due to fine-grained inter-species variation and high intra-species variation. This paper presents our approach for the FungiCLEF 2025 competition, which focuses on few-shot fine-grained visual categorization (FGVC) using the FungiTastic Few-Shot dataset. Our team (DS@GT) experimented with multiple vision transformer models, data augmentation, weighted sampling, and incorporating textual information. We also explored generative AI models for zero-shot classification using structured prompting but found them to significantly underperform relative to vision-based models. 
    Our final model outperformed both competition baselines and highlighted the effectiveness of domain-specific pretraining and balanced sampling strategies. Our approach ranked 35/74 on the private test set in post-completion evaluation, this suggests additional work can be done on metadata selection and domain-adapted multi-modal learning. Our code is available at \url{https://github.com/dsgt-arc/fungiclef-2025}.
\end{abstract}


\maketitle

% We recommend splitting your main document into smaller parts that are easier to navigate.
% The input command "imports" the contents of the file into the current location.
% Prefixing the document with a number allows for natural string sorts.
\input{sections/00_main}

\bibliography{main}
\end{document}

%% file: sections/00_main.tex
\section{Introduction}
Accurate identification of fungi species is challenging and often requires microscopic examination \cite{Unambiguous-identification-of-fungi}, which places this task beyond simple image classification and falls into fine-grained visual categorization (FGVC). This paper documents and evaluates our approaches as part of the FungiCLEF 2025 \cite{fungi-clef-2025} @ CVPR-FGVC \& LifeCLEF 2025 \cite{lifeclef2025} competition. The goal of this competition is fungi species recognition and the evaluation metric is top-5 accuracy.

\begin{figure}[h!]
    \centering
    \includegraphics[width=0.25\linewidth]{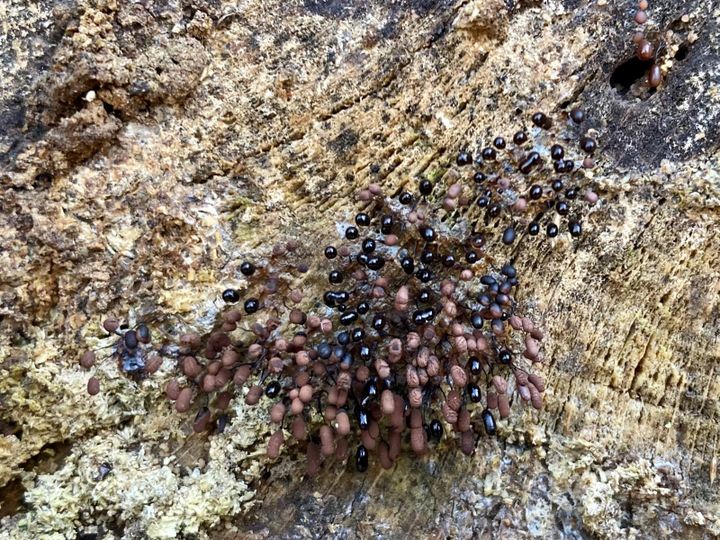}
    \includegraphics[width=0.25\linewidth]{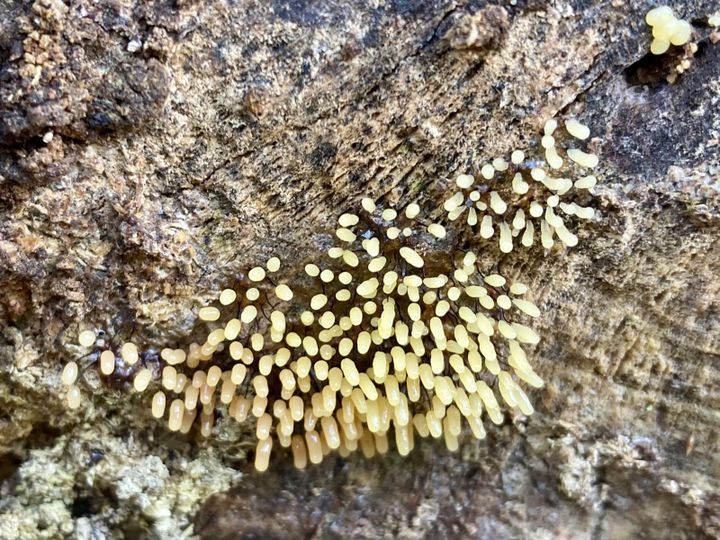}
    \includegraphics[width=0.25\linewidth]{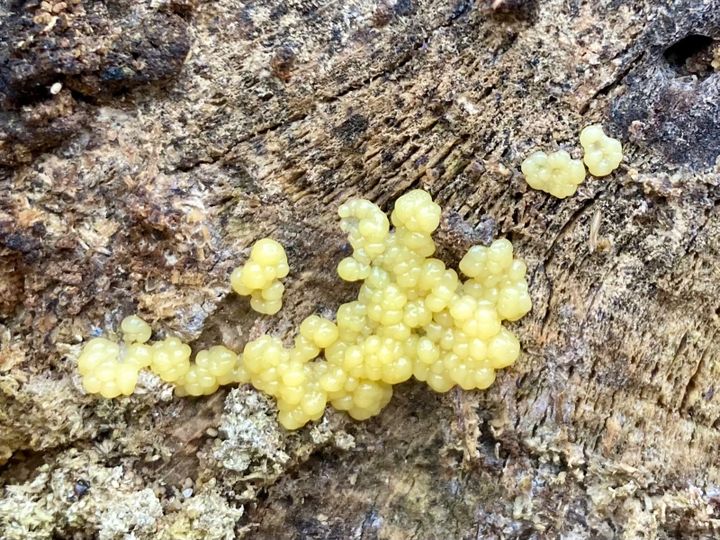}
    \caption{Example of intra-species variation - Species: Comatricha alta}
    \label{fig:intra}
\end{figure}

\begin{figure}[h!]
    \centering
    \includegraphics[width=0.2\linewidth]{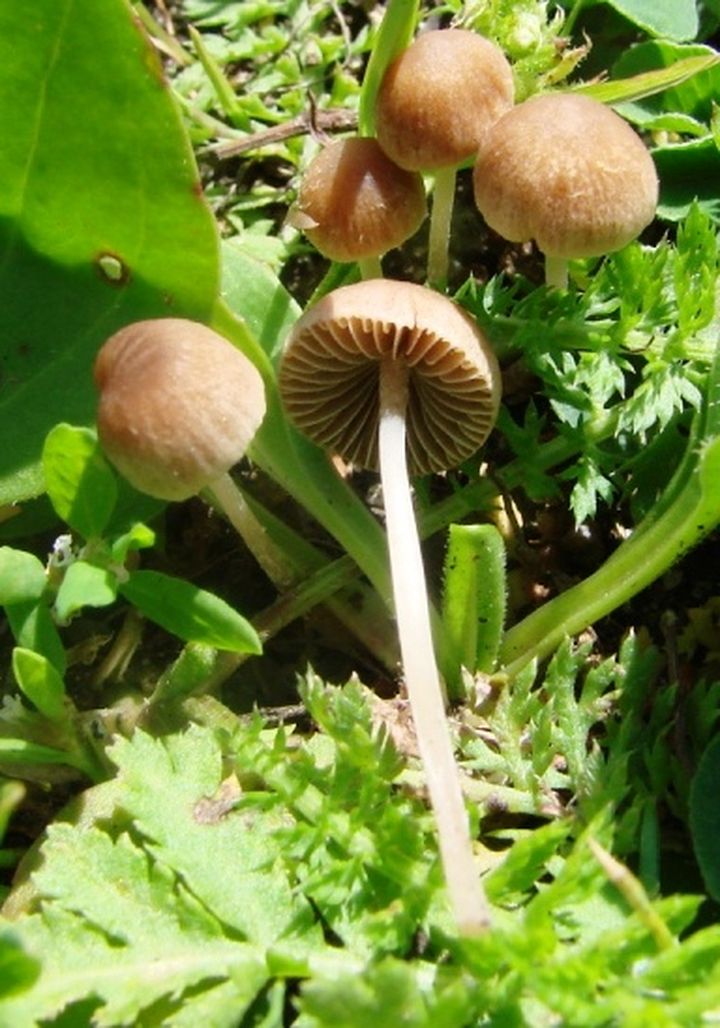}
    \includegraphics[width=0.25\linewidth]{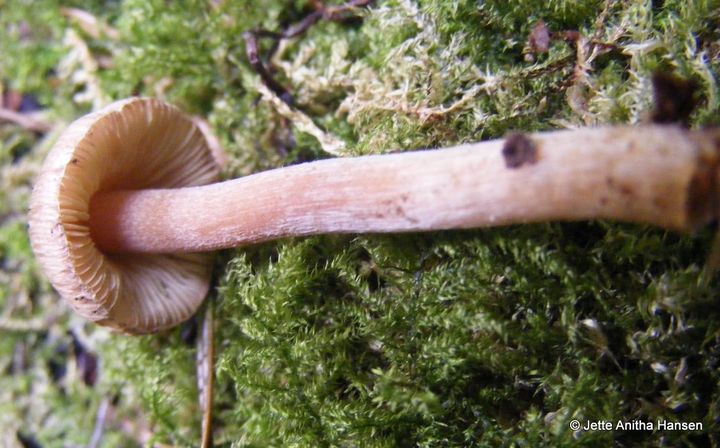}
    \includegraphics[width=0.25\linewidth]{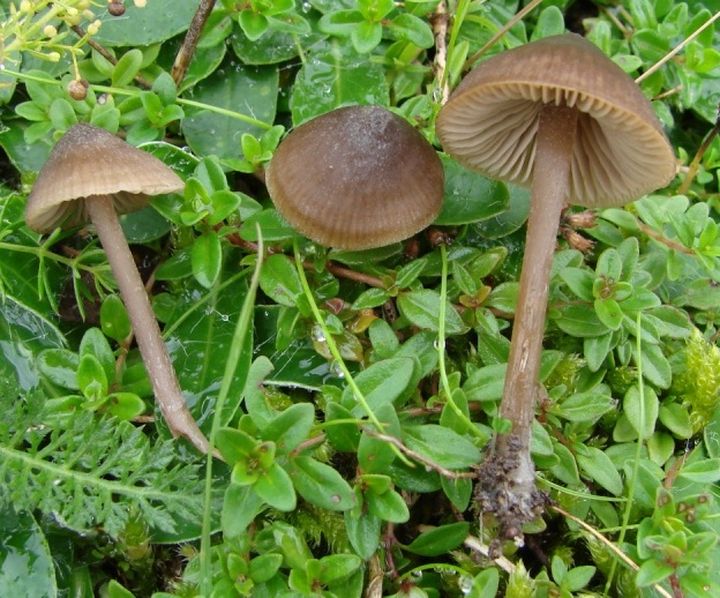}
    \caption{Example of inter-species similarity and variation from three distinct families. Species left to right: Psathyrella citerinii, Inocybe assimilata, and Entoloma favrei }
    \label{fig:inter}
\end{figure}

Multiple aspects make this task challenging. First, there is intra-species variation due to factors such as genotype, local conditions, time of year, and age (Figure \ref{fig:intra}) \cite{Overview-of-FungiCLEF-2024}. Second, there is subtle inter-class variation even at higher order taxonomic ranks, which makes it difficult for pattern learning algorithms to differentiate (Figure \ref{fig:inter}) \cite{Overview-of-FungiCLEF-2024}. 

\begin{figure}[h!]
    \centering
    \includegraphics[width=1\linewidth]{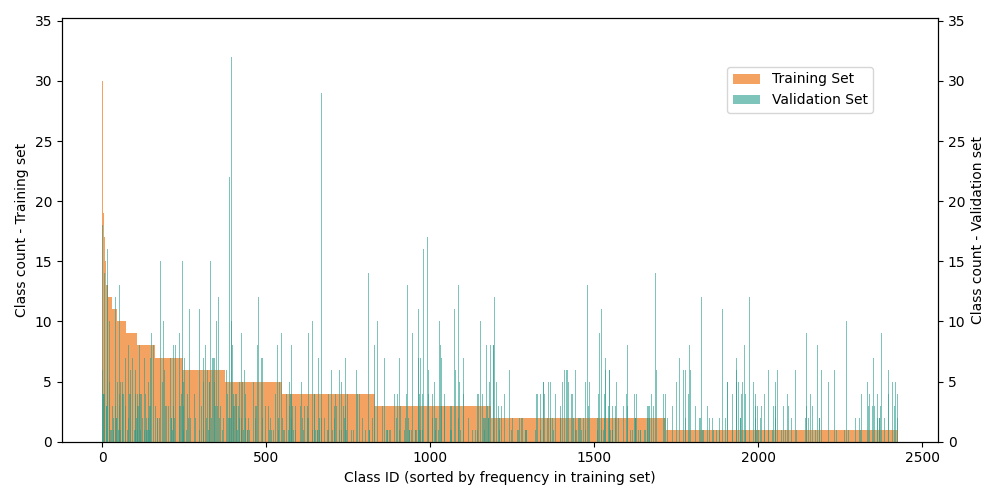}
    \caption{Distribution of classes in training and validation datasets }
    \label{fig:class-balance}
\end{figure}
\subsection{Dataset Overview}
The dataset provided for the competition is the few-shot subset of the FungiTastic dataset, a collection of fungal records continuously collected over a twenty-year span \cite{FungiTastic}. Each observation in the dataset contains associated images, metadata, and vision language model (VLM), Molmo \cite{molmo}, generated caption. The metadata contains information such as date, location, substrate, and full taxonomic ranks. It is important to note that the task is to classify images based on category\_id, which has a slightly different count than species. The training dataset contains 7,819 images with 2,413 unique species and 2,427 unique category\_id. The validation dataset contains 2,285 images with 569 species and 570 unique category\_id. The test dataset contains 1,911 images with no taxonomic ranks. The provided image dataset contains the training, validation, and testing sub-datasets. Each sub-dataset contains images in different maximum pixel sizes, ranging from 300p to full-size images.

The datasets do not have the same category\_id distribution (Figure \ref{fig:class-balance}). In the chart, the category\_id is mapped to class ID and then sorted by frequency, with category\_id 2383 appearing most frequently. Both datasets exhibit class imbalance, with the most common class having approximately 30 images and multiple classes having only 1 image. 

\section{Related Work}
Previous work by the DS@GT group for FungiCLEF 2024 demonstrated the strong performance of using DINOv2 vision transformers in image classification \cite{fungiclef2024}. Last year's winner, Team IES, combined image embeddings from Swin Transformer V2 \cite{swin} with metadata features from multi-layer-perception for species classification \cite{ies}.

\section{Methodology}
Our benchmark approach uses PlantCLEF 2024 \cite{plantclef} embeddings, weighted sampling \cite{sampler}, and Mixup \cite{mixup}. Using off-the-shelf generative AI models, multi-modal approach of combining text embeddings with image embeddings, and multi-objective loss were also explored. The competition evaluation metric is top-k accuracy, with k = 5.
\begin{equation}
\text{Top-k Accuracy} = \frac{1}{N} \sum_{i=1}^{N} 1\left[y_i \in \hat{Y}_i^{(k)}\right]
\end{equation}

The cloud computing resources were funded by the Data Science at Georgia Tech (DS@GT). Data and computing was hosted by the Partnership for an Advanced Computing Environment (PACE) at Georgia Tech \cite{PACE}. 

\subsection{Benchmark Methodology}
Our benchmark methodology can be summarized in a few steps:
\begin{enumerate}
    \item Image embeddings from PlantCLEF 2024 model
    \item Weighted sampling to balance the training dataset
    \item Mixup on batches during training
    \item Linear classifier
    \item Mixup loss with cross-entropy \cite{crossentropy}
\end{enumerate}
\begin{figure}[h!]
    \centering
    \includegraphics[width=1\linewidth]{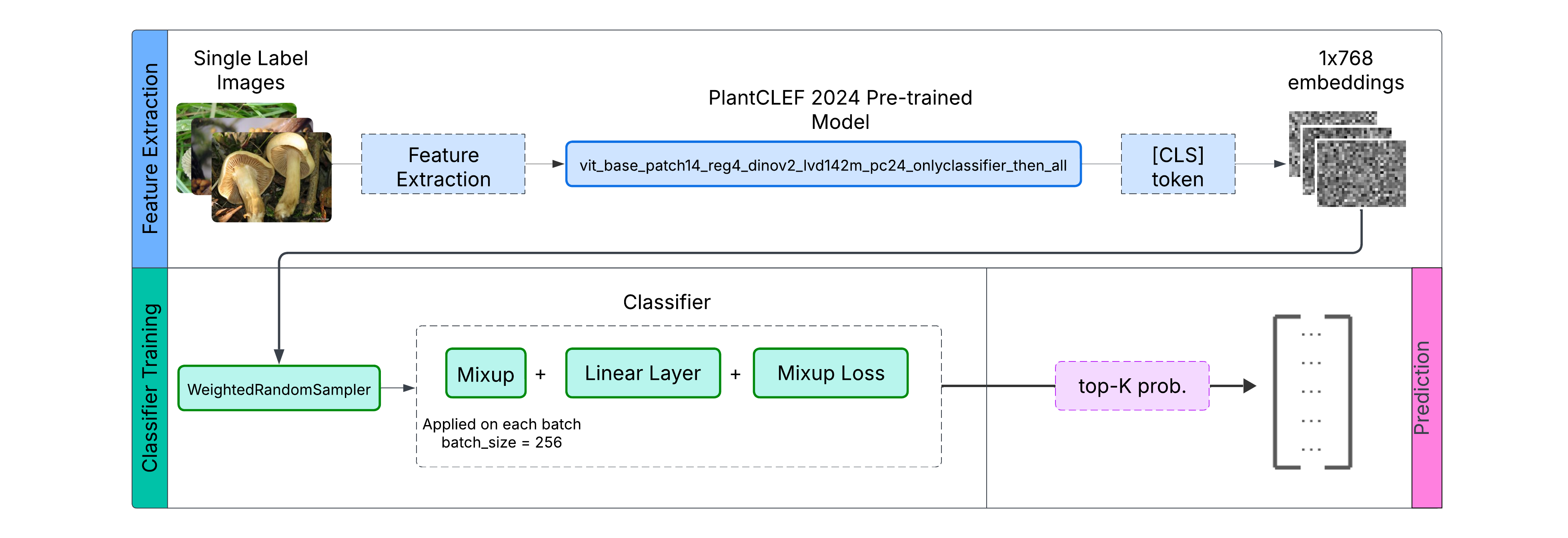}
    \caption{Pipeline of the benchmark methodology }
    \label{fig:pipeline}
\end{figure}

\subsubsection{Dataset Preparation}
We pre-computed both the image and text embeddings and stored them in parquet files for a modular experimentation workflow. 
We encountered "Premature End of JPEG file" error when reading images because some images did not end with the default hex code. This may result in unintended side effects during training. This error was solved by loading the image with OpenCV and then saving it again \cite{fix-end-of-file}. There was one corrupted image in the validation 720p set; we did not use this image since we only used the full-size images for our pipeline. 

All approaches except Generative AI involved using the pytorch lightning library in training the classifier. The hyperparameters used are as follows: batch size of 256, 50 maximum epochs with early stopping of 3, Adam \cite{adam} optimizer, learning rate of $5 \cdot 10^{-4}$ and no learning rate scheduler. 

\subsubsection{Image Embeddings}
We experimented with multiple transformer models: Facebook DINOv2 \cite{dinov2}, PlantCLEF 2024 pre-trained model\cite{plantclef}, FungiTastic BEiT \cite{FungiTastic}, and FungiTastic ViT\cite{FungiTastic}. A summary of the models used is shown in Table \ref{tab:transformer-list} and Table \ref{tab:transformer-detail}.

DINOv2 was selected for its state-of-the-art performance in computer vision tasks and its strong results in FungiCLEF 2024\cite{fungiclef2024}. The PlantCLEF model was selected due to its foundation in DINOv2 and its pre-training on 1.4 million plant images from the Pl@ntNet database, offering the potential benefits of transfer learning. The two FungiTastic models were selected because they were pre-trained on the fungi images. 

The image embeddings from the PlantCLEF 2024 model used in our benchmark methodology have a size of 768.

\begin{table}[h!]
    \caption{Summary of models explored for image embeddings generation}
    \centering
    \begin{tabular}{ccc}
        \toprule
         \textbf{Shorthand}& \textbf{Model Name}\\
         \midrule
        DINOv2 & dinov2-base \cite{dinov2} \\
        PlantCLEF 2024 & vit\_base\_patch14\_reg4\_dinov2\_lvd142m\_pc24\_onlyclassifier\_then\_all \cite{plantclef}\\
        FungiTastic BEiT & beit\_base\_patch16\_224.in1k\_ft\_fungitastic\_224\cite{FungiTastic}\\
        FungiTastic ViT & vit\_base\_patch16\_224.in1k\_ft\_fungitastic\_224 \cite{FungiTastic}\\
        \bottomrule
    \end{tabular}
    \label{tab:transformer-list}
\end{table}

\begin{table}[h!]
    \caption{Details of models explored for image embeddings generation}
    \centering
    \begin{tabular}{cccc}
        \toprule
         \textbf{Shorthand} & \textbf{Architecture} & \textbf{Hidden Size} & \textbf{Pretraining/Finetuning}\\
         \midrule
        DINOv2 & ViT-B/14 \cite{dinov2}  & 768 & LVD-142M/None\\
        PlantCLEF 2024 & ViT-B/14 \cite{dinov2} & 768 & DINOv2 on LVD-142M/Pl@ntNet \\
        FungiTastic BEiT & BEiT-B/16\cite{BEIT} & 768 & ImageNet-1k/FungiTastic\\
        FungiTastic ViT & ViT-B/16 \cite{vit16}  & 768 & ImageNet-1k/FungiTastic\\

        \bottomrule
    \end{tabular}
    \label{tab:transformer-detail}
\end{table}

\subsubsection{Weighted Sampling and Mixup}
To mitigate the effects of class imbalance observed in Figure \ref{fig:class-balance}, we experimented with the PyTorch's  \texttt{WeightedRandomSampler} \cite{sampler} using weights calculated with inverse class frequency via \texttt{compute\_sample\_weight} from the \texttt{sklearn} library, on the training dataset. This sampling strategy was implemented in the data loader to ensure that minority classes were sampled more frequently during training.

We also experimented with Mixup \cite{mixup} to increase the influence of minority classes. Mixup was implemented in the classifier and applied to batches provided by the data loader.
Mixup encourages the classifier model to generalize better by interpolating features and labels between classes. 
In our implementation, the embeddings extracted from the training dataset were linearly combined with a shuffled version to generate an augmented set using the equations:

\begin{equation}
\begin{aligned}
\tilde{x} &= \lambda x_i + (1 - \lambda) x_j \\
\end{aligned}
\end{equation}
\begin{equation}
\mathcal{L}_{\text{Mixup}} = \lambda \cdot \mathcal{L}(f(\tilde{x}), y_i) + (1 - \lambda) \cdot \mathcal{L}(f(\tilde{x}), y_j)
\end{equation}

where \( \lambda \sim \text{Beta}(\alpha, \alpha) \), \( x \) denotes image embeddings, \( y \) denotes the label targets, \( i \) indexes the original mini-batch, and \( j \) indexes a randomly shuffled version of the same mini-batch. In the competition approach, \( \alpha = 2.0 \), 256 batch size, and 10 epochs were used to evaluate the impact of Mixup. An \( \alpha = 2.0 \) was inspired by Manifold Mixup \cite{manifold} to encourage an greater generalization due to the small dataset. 

\begin{figure}[h!]
    \centering
    \includegraphics[width=1\linewidth]{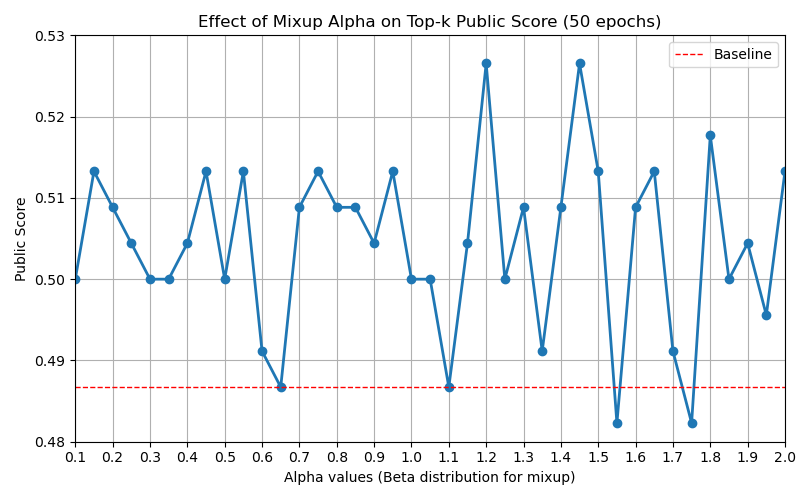}
    \caption{Effect of Mixup Alpha on Top-K Accuracy}
    \label{fig:alpha}
\end{figure}

In our post-competition evaluation, we increased the epochs used in the Mixup \cite{mixup} only approach to 50 to make its results more comparable with the other approaches discussed in Section 4 Results as all other approaches used 50 max epochs. Here, we evaluated the results using \( \alpha\) values ranging from [0.1, 2.0], which encompasses the recommended ranges from the Mixup \cite{mixup} paper and the Manifold Mixup \cite{manifold} paper (Figure \ref{fig:alpha}).  An \( \alpha = 1.20 \) and \( \alpha = 1.45 \) achieved the highest public score and these two values were used to run additional experiments with a finetuned Mixup and weighted sampling.

\subsection{Additional Methodologies}
\subsubsection{Text Embeddings}
We used ModernBERT-Large\cite{modernBERT}, a state-of-the-art BERT variant optimized for efficiency, to compute 1024-dimensional text embeddings. We concatenated text from categories present in the test metadata file with the generated captions to form a single string. The results were saved in a parquet file.

In post-competition evaluation, we also used BioBERT-Large \cite{biobert}, a domain specific BERT based model pre-trained on biomedical corpora to compute 1024-dimensional text embeddings. There is a potential for transfer learning between biomedical texts and fungi textual information since both fall under the domain of biology. Again, we concatenated text from categories present in the test metadata file with the generated captions to form a single string.

In a multi-modal classifier, the image and text embeddings are fed through their own linear layers with the same output size of 256. The image and text embeddings are then concatenated, normalized, and fed into a 512 input size linear classification layer.

\subsubsection{Multi-Objective Loss GradNorm}

Inspired by the evaluation metrics used in FungiCLEF 2024\cite{fungiclef2024}, we experimented with a multi-objective classification framework to jointly predict category\_id, poisonous, genus, and species. 
Each objective has its own classification head and loss function: cross-entropy loss for category\_id, genus, and species, and binary cross-entropy loss for poisonous. 
To prevent a single objective from dominating classification and to encourage balanced learning, we implemented GradNorm \cite{gradnorm}.
GradNorm allows dynamically assigning weights for each objective when calculating loss. 
We introduced a learnable weight for each objective and computed the gradient norms with respect to the shared parameters.

\subsubsection{Generative AI}

We explored the use of generative AI techniques to predict species in the dataset.
Many commercially available multi-modal large language models are vision-language models, where vision and language modalities are fused through an attention mechanism.
We implement a zero-shot prompting method across three API providers using the OpenRouter platform and leverage structured output to enforce the structural regularity of the results.

\begin{table}[htbp]
    \centering
    \caption{Models used for experiments via OpenRouter.}
    \label{tab:llm_models}
    \begin{tabular}{@{}l c r r r l@{}}
        \toprule
        \textbf{Model Name} & \textbf{Release Date} & \textbf{Context} & \textbf{Input} & \textbf{Output} & \textbf{Vision Input} \\
        & & \textbf{(tokens)} & \textbf{(\$/M)} & \textbf{(\$/M)} & \textbf{(\$/K images)} \\
        \midrule
        google/gemini-2.0-flash-001 & 2025-02-05 & 1,048,576 & \$0.10 & \$0.40 & \$0.026 \\
        openai/gpt-4.1-mini-2025-04-14 & 2025-04-14 & 1,047,576 & \$0.40 & \$1.60 & N/A \\
        google/gemini-2.5-flash-preview-04-17 & 2025-04-17 & 1,048,576 & \$0.15 & \$0.60 & \$0.619 \\
        mistralai/mistral-medium-3 & 2025-05-07 & 131,072 & \$0.40 & \$2.00 & N/A \\
        \bottomrule
    \end{tabular}
\end{table}

We perform three rounds of prompting across family, genus, and species per test image to logarithmically reduce the search space and ensure that only species within the training set are used.
Each round of prompting relies on the prompt used in listing~\ref{lst:prompt}.
We append a yaml list of all the candidate items to rank.
We append all available images for a single image ID (which can range from one to a dozen images) as context to the completion.
We request a list of 20 ranked candidates, including an item name and a corresponding confidence score.
The results are validated against the candidate list and accepted if at least half of the results are valid, i.e., there exists an item that is within 90\% of the string by normalized edit distance. 
For human debugging purposes, we also have the LLM generate a reason for the decision.

\begin{tcolorbox}[
    breakable, % Allows this tcolorbox to break across pages
    colframe=black,       % Example frame color
    boxrule=0.5pt,        % Thickness of the frame
]
\begin{lstlisting}[
    basicstyle=\ttfamily\footnotesize,
    breaklines=true,
    label={lst:prompt}
]
Accurately identify and assign the correct {class_type} label to each image of
fungi, protozoa, or chromista utilizing all provided image views and associated
metadata (location, substrate, season) to ensure precision, especially for
fine-grained distinctions. Choose the top twenty most relevant labels ranked in
order from the available class labels, a confidence on the Likert scale between
1-5 on not-confident to confident and provide short reasoning (in under 50
words) for your selection.
\end{lstlisting}
\end{tcolorbox}

In the first round of prompting, we provide a list of all families and ask the LLM to rank the top 20 species relevant to the test images.
We use the most relevant families to generate a candidate list of genera. 
We then use this to generate a candidate list of species.
We provide the top 10 species as the final result of the competition.

\section{Results}
\subsection{Image Embeddings Results}
The best performing models to pre-compute the image embeddings were the PlantCLEF 2024 \cite{plantclef} model and the FungiTastic ViT \cite{FungiTastic} model. The embeddings from each model were passed into a linear layer to generate the predictions. The top-5 accuracy public score is then used to select the model to use in our benchmark methodology (Table \ref{tab:model-study}). PlantCLEF 2024 \cite{plantclef} was selected as our baseline classifier and incorporated into our best performing approach. 
\begin{table}[h!]
    \caption{Performance of models used for image embeddings}
    \centering
    \begin{tabular}{ccc}
        \toprule
         & Top-5 Acc., Public (\%) & Difference (\%) \\
         \midrule
        PlantCLEF 2024 & \textbf{48.672} & -\\
        DINOv2 & 47.345 & -1.327\\
        FungiTastic BEiT & 42.477 & -6.195\\
        FungiTastic ViT & \textbf{48.672} & 0\\
        \bottomrule
    \end{tabular}
    \label{tab:model-study}
\end{table}
\subsection{Ablation Study}
The results from our various approaches are compiled in Table \ref{tab:ablation}. Our best in-competition approach was with Mixup \cite{mixup} \( \alpha = 2.0 \) and weighted sampling with a private top-5 accuracy score of 40.75. In post-competition evaluation, a finetuned \( \alpha = 1.20 \) achieved the highest private score when combined with weighted sampling and \( \alpha = 1.45 \) achieved the highest private score when used by itself. 

We found that Mixup with a tuned \( \alpha\) is the single technique with the greatest positive impact with an increase of 4.27\% on the private score. Weighted sampling provided a much smaller increase in accuracy on its own and had minimal effect when combined with a tuned \( \alpha\). Lastly, we find that incorporating metadata + caption and a multi-objective GradNorm \cite{gradnorm} approach of classifying category\_id, poisonous, species, and genus had a negative impact on the prediction accuracy.

\begin{table}[h!]
    \caption{Ablation study of our different approaches}
    \centering
    \setlength{\tabcolsep}{4pt} % reduce spacing between columns
    \begin{tabular}{@{}ccc@{}}
        \toprule
        \textbf{Classification method} & \multicolumn{2}{c}{\textbf{Top-5 Acc.}} \\
        \cmidrule(lr){2-3}
         & \textbf{Public (\%)} & \textbf{Private (\%)} \\
         \midrule
        PlantCLEF 2024 Model Baseline & 48.672 & 43.078\\
        w/ weighted sampling & 50.442 & 44.372\\
        w/ Mixup (\(\alpha\) = 2.00)* (Competition) & 46.460 & 40.750\\
        w/ Mixup (\(\alpha\) = 1.20) (Post-Comp) & 52.654 & 44.760\\
        w/ Mixup (\(\alpha\) = 1.45) (Post-Comp) & 52.654 & \textbf{47.347}\\
        w/ Mixup (\(\alpha\) = 2.00) \& weighted sampling (Competition) & \textbf{49.557} & \textbf{45.407}\\
        w/ Mixup (\(\alpha\) = 1.20) \& weighted sampling (Post-Comp) & 50.884 & \textbf{46.830}\\
        w/ Mixup (\(\alpha\) = 1.45) \& weighted sampling (Post-Comp) & \textbf{53.982} & 46.054\\
        w/ text (ModernBert) (Competition) & 46.460 & 38.421\\
        w/ text (BioBert) (Post-Competition) & 45.132 & 40.620\\
        w/ GradNorm and weighted sampling & 42.920 & 39.197\\
        w/ text, GradNorm, and weighted sampling & 39.823 & 36.093\\
        \midrule
        Gemini 2.0 Flash & 14.159 & 12.548\\
        Gemini 2.5 Flash & 11.504 & 13.583\\
        OpenAI GPT 4.1 Mini & 5.309 & 6.209\\
        MistralAI Mistral Medium 3 & 4.424 & 3.104\\
        \bottomrule
        * Trained with 10 epochs. All other classifier approaches are trained with 50 epochs
    \end{tabular}
    \label{tab:ablation}
\end{table}

\subsection{Leaderboard Results}

Our team's result beat both competition baselines: BioCLIP \cite{bioclip} + FAISS + Prototypes and BioCLIP + FAISS + NN. However, our result falls short of the leaders in the competition. We are ranked 37/74 on the public leaderboard and 35/74 on the private leaderboard. These results are summarized in Table \ref{tab:leaderboard}

\begin{table}[h!]
    \caption{Public and Private test set scores of the leaderboard}
    \centering
    \begin{tabular}{ccc}
        \toprule
         &\textbf{ Name}  & \textbf{Top-5 Acc. (\%)} \\
        \midrule
        Public & PlantCLEF + Mixup (\(\alpha\) = 2.00) + Weighted Sampling (Competition) & 49.557\\
        Private & PlantCLEF + Mixup \(\alpha\) = 2.00) + Weighted Sampling (Competition) & 45.407\\
        \midrule
        Public & PlantCLEF + Mixup (\(\alpha\) = 1.20) + Weighted Sampling (Post-Comp) & 50.884\\
        Private & PlantCLEF + Mixup (\(\alpha\) = 1.20) + Weighted Sampling (Post-Comp) & 46.830\\
        \midrule
        Public & Rank 1 - Jack Etheredge & \textbf{81.858}\\
        Private & Rank 1 - Jack Etheredge & \textbf{78.913}\\
        \midrule
        Public & Rank 2 - Hasan Oetken & 80.530\\
        Private & Rank 2 - hard\_work & 78.137\\
        \midrule
        Public & BioCLIP + FAISS + Prototypes Baseline & 33.185\\
        Private & BioCLIP + FAISS + Prototypes Baseline & 26.649\\
        \midrule
        Public & BioCLIP + FAISS + NN Baseline & 28.318\\
        Private & BioCLIP + FAISS + NN Baseline & 24.708\\
        \bottomrule
    \end{tabular}
    \label{tab:leaderboard}
\end{table}

\subsection{Validation Dataset Performance}
\begin{figure}[h!]
    \centering
    \includegraphics[width=1\linewidth]{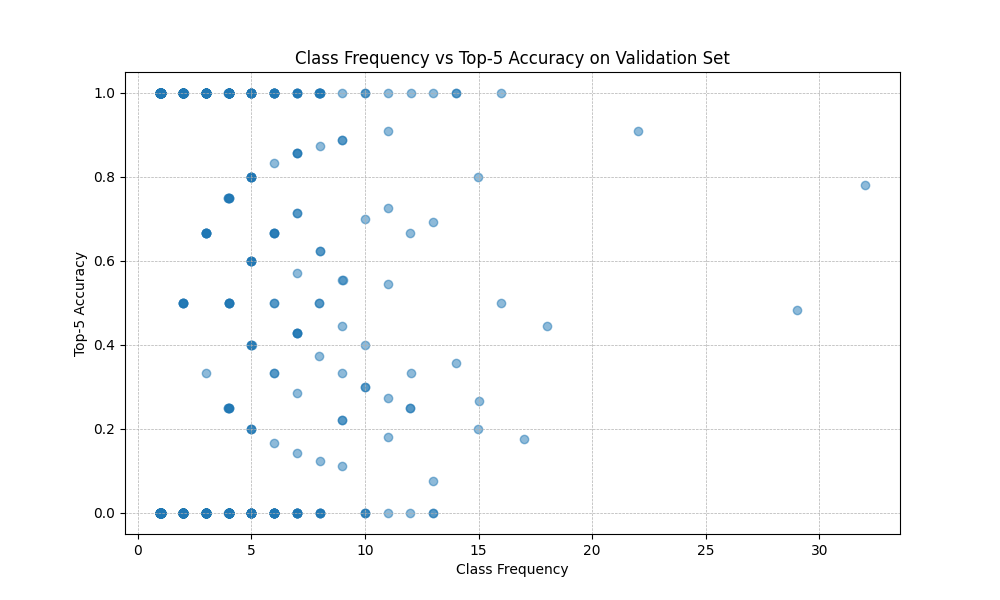}
    \caption{Class Frequency vs Top-5 Accuracy on Validation Set}
    \label{fig:validateperformance}
\end{figure}

In Figure \ref{fig:validateperformance}, we plot the class frequency versus the top-5 accuracy on the validation dataset. The class frequency is calculated on a per image basis, and not a per observation basis. There can be multiple images under an observation. The concentration of points at accuracy 1.0 and 0.0 (shown as darker points) at the rarer classes shows that the classifier often achieves perfect or zero Top-5 accuracy due to the small sample size. This highlights the volatility in the classification accuracy in class imbalanced datasets.  

\section{Discussion}
\subsection{Weighted Sampling and Mixup}
As seen in Table \ref{tab:ablation}, Mixup \cite{mixup} with a tuned \(\alpha = 1.20\) and \(\alpha = 1.45\) had the greatest positive impact on our baseline accuracy. This is different from the recommended range of [0.1,0.4] for \(\alpha\) suggested by the Mixup paper and could be because the Mixup is applied at the feature level instead at the raw inputs. Applying Mixup at the feature level is closer to the Manifold Mixup \cite{manifold} approach which applies Mixup between intermediate layers of a neural network and uses \(\alpha = 2.00\). From Figure \ref{fig:alpha}, there is no clear monotonic trend as \(\alpha\) changes, however all values of \(\alpha\) except for two resulted in an improvement over the baseline. The lack of a monotonic trend may suggest adding more learnable layers to our classifier is needed to flatten class boundaries and reduce volatility as seen in Manifold Mixup \cite{manifold}. 

Weighted sampling \cite{sampler} was another approach that had a positive impact on our baseline accuracy, albeit to a smaller extent than a tuned Mixup. The modest increase in accuracy indicates that while it helps the model see rare classes more often, it alone does not sufficiently address the challenges of learning robust patterns for underrepresented classes. The mixed results when combined with Mixup shows that there is diminishing returns in applying multiple sampling approaches. 

\subsection{Image Embeddings Generation}

Among the different models evaluated for image embedding generation, the PlantCLEF 2024 and FungiTastic ViT models performed the best (Table \ref{tab:model-study}. These two models slightly edged out general use DINOV2. Although the improvement is small, this suggests that domain-adapted models can offer an advantage over general use models in few-shot fine grained species classification. This finding is similar to the few-shot results presented in the FungiTastic paper \cite{FungiTastic}, in which BioCLIP\cite{bioclip} outperformed DINOv2 and CLIP\cite{clip}. 

\subsection{Text Embeddings}
The inclusion of metadata and captions had a negative impact on our classifier performance. This is contrary to the findings presented in the FungiTastic paper \cite{FungiTastic}, where the incorporation of metadata did improve performance. This discrepancy is likely due to our inclusion of extraneous or weakly informative metadata such as \texttt{district}, \texttt{countryCode}, and \texttt{hasCoordinate}.  

\subsection{Generative AI Results}

Current-generation multi-modal LLMs are not effective at generalizing to the domain-specific task of labeling fungus images, at least within our price range.
Our best model is from the Gemini family of models, scoring around 13\% on the private leaderboard.
Our choice of models is dictated by cost. 
For example, Gemini Pro is about 10 times as expensive as the Flash series of models. 
Gemini Flash was at a level of cost-effectiveness that we were willing to experiment with, and initial experimentation led us to hold off on trying models with higher token usage, which is generally associated with "thinking" or "reasoning" capabilities.
We then chose GPT-4.1-mini and Mistral as models that had both structured output and image inputs.
The set of models that accepted both of these constraints is much smaller than we would have liked and precluded models such as Anthropic Claude.
We summarize the costs associated with this approach in table~\ref{tab:llm_usage_summary}, which comes close to a total of \$30 over 15k requests.

\begin{table}[htbp]
    \centering
    \caption{Model Usage and Cost Summary.}
    \label{tab:llm_usage_summary}
    \begin{tabular}{@{}l rrr@{}}
        \toprule
        \textbf{Model Name} & \textbf{Total Tokens (M)} & \textbf{Total Requests (K)} & \textbf{Total Cost (\$)} \\
        \midrule
        mistralai/mistral-medium-3 & 25.20 & 3.60 & 13.80 \\
        openai/gpt-4.1\_mini-2025-04-14 & 28.10 & 3.04 & 12.70 \\
        google/gemini-2.5\_flash-preview-04-17 & 7.84 & 3.65 & 2.29 \\
        google/gemini-2.0\_flash-001 & 8.36 & 3.99 & 2.06 \\
        \bottomrule
    \end{tabular}
\end{table}

Note that while we limited models to structured outputs, we can simulate this in a two-pass methodology, where a stronger model generates results in a particular semi-structured shape, and a second, smaller but cheaper model converts this into a structured output via JSON Schema.
However, this requires more boilerplate code and effort than we were willing to explore at this point, given the performance relative to stronger vision-first approaches.
We also note that our three-round approach was necessary because there are limits to the structured schema API.
For example, one of the first things we tried was to return a list of strings where a string must be part of a particular enumeration.
However, enumerations are supported only up to a certain number of elements, which are undocumented, if supported at all.
Another reason is that the context window significantly influences which elements are recalled from the list of available class elements.
If we were to include all species in one big list, there is a good chance that not every species would be considered from that list due to limitations of context locality.
This behavior is challenging to describe quantitatively due to the accelerated pace of development of these models in production and the associated cost of running experiments.

We also note a few limitations in our methodology.
First, LLMs are strongly affected by the amount of stochasticity introduced at token generation time (i.e., temperature).
As such, the runs of our algorithm will change significantly over time, making it challenging to reproduce our results exactly.
However, there are two approaches to mitigate reproduction issues, given that the cost of a single Gemini test run is about \$2. 
The first option is to lower the temperature of the model, which is often supported.
Another approach is to run several iterations of a model and aggregate the final results.
The ideal solution would take on a Monte-Carlo tree search flavor, where we would sample the top-k elements many times and produce some probabilistic taxonomic tree based on knowledge embedded in the LLM.

FungiCLEF is a domain-specific task that is relatively resource-poor relative to the general task of information recall from large pools of publicly available text.
However, it is impressive that these models can get any results at all.
It would be interesting to gain a deeper understanding of the vision-question-based capabilities of these models, perhaps by using a smaller subset that considers the general challenges of the fungi dataset while managing costs.
What might make the most sense is fine-tuning a smaller VLM, such as Gemma \cite{team2024gemma}, Phi \cite{abdin2024phi}, or Llama \cite{touvron2023llama}, on the FungiCLEF dataset and seeing whether these smaller models can be effectively tuned for domain-specific language queries.

\section{Future Work}
Improvements can be made to address to disparity in class distributions between the training and validation datasets as observed in Figure \ref{fig:class-balance}. One approach, seen in previous research, is to combine the provided datasets \cite{fungiclef2024}, then we can resplit them to have similar distributions. 
In addition, improvements could be made to the selection and processing of textual data. Rather than incorporating all available metadata fields, future work should prioritize informative features. As a starting point, we propose using the three metadata attributes highlighted in the FungiTastic paper \cite{FungiTastic}, which have been shown to be effective in improving model performance. In addition, more learnable layers can be added to the classifier with Mixup \cite{mixup} applied at a random layer to more closely follow the approach proposed in Manifold Mixup \cite{manifold}.
Continuing with our generative AI approach, one can experiment with costlier models, or implement a Monte-Carlo tree search as discussed in Section 5.4.

\section{Conclusions}
In this paper, we present our approach to tackle the challenge of FungiCLEF 2025 few-shot fine-grained visual classification (FGVC) using vision transformer embeddings. We explored a range of models, such as DINOv2, PlantCLEF 2024, and FungiTastic pre-trained models, ultimately selecting the PlantCLEF 2024 model for our benchmark approach due to its strong performance and transfer learning. 
To mitigate the class imbalance in the dataset, we implemented weighted sampling and Mixup, with Mixup providing the most significant performance gain. We also experimented with incorporating textual metadata and multi-objective learning GradNorm, but found these approaches to be detrimental, likely due to noisy or weakly informative inputs.
Our final competition and post-competition classifiers outperformed both competition baselines and demonstrated the importance of domain-specific embeddings and balancing strategies. However, a significant performance gap with the leaders in the competition indicates the need for further exploration of alternate classifier architectures and improved metadata integration. 

\section*{Acknowledgements}

We thank the Data Science at Georgia Tech (DS@GT) CLEF competition group for their support.
This research was supported in part through research cyberinfrastructure resources and services provided by the Partnership for an Advanced Computing Environment (PACE) at the Georgia Institute of Technology, Atlanta, Georgia, USA \cite{PACE}. 

\section{Declaration on Generative AI}
During the preparation of this work, the authors used Chat-GPT-4 in order to: drafting content, Grammar and spelling check. After using these tool(s)/service(s), the authors reviewed and edited the content as needed and take full responsibility for the publication’s content.